\documentclass[11pt]{article}
\usepackage{nodalida2021}
\usepackage{times}
\usepackage{url}

\usepackage{breakurl}
\usepackage[hidelinks,breaklinks]{hyperref}
\usepackage{latexsym}
\usepackage{booktabs, multirow}
\usepackage{comments}
\usepackage{arydshln, balance}
\definecolor{asparagus}{rgb}{0.53, 0.66, 0.42}
\definecolor{bluebell}{rgb}{0.64, 0.64, 0.82}
\newauthor[SH]{Simon}{asparagus}   
\newauthor[NT]{Nina}{bluebell}  
\aclfinalcopy 

\title{SuperSim: a test set for word similarity and relatedness in Swedish}

\author{Simon Hengchen, Nina Tahmasebi\\
Språkbanken Text, Department of Swedish\\
University of Gothenburg\\
{{\tt \{simon.hengchen;nina.tahmasebi\}@gu.se}} \\}

\date{}

\begin{document}
\maketitle
\begin{abstract}
Language models are notoriously difficult to evaluate. 
We release SuperSim, a large-scale similarity and relatedness test set for Swedish built with expert human judgments. The test set is composed of 1,360 word-pairs 
independently judged for both relatedness and similarity by five annotators. We evaluate three different models (Word2Vec, fastText, and GloVe) trained on two separate Swedish datasets, namely the Swedish Gigaword corpus and a Swedish Wikipedia dump, to provide a baseline for future comparison. 
We release the fully annotated test set, code, baseline models, and data.\footnote{\url{https://zenodo.org/record/4660084}.}
\end{abstract}

\section{Introduction}

It is said that a \textit{cup} and \textit{coffee} are not very similar while \textit{car} and \textit{train} are much more so given that they share multiple similar features. Instead, \textit{cup} and \textit{coffee} are highly related, as we typically enjoy the one in the other. Of course, an immediate question that arises is whether we have words that are similar but not related? Existing similarity datasets have tended to rate words for their similarity, relatedness, or a mixture of both, but not either or. However, without both kind of information, we cannot know if words are related but not similar, or similar but not related. 

The most common motivation for using word similarity datasets, such as SimLex-999 \citep{hill-etal-2015-simlex} and WordSim353 \citep{finkelstein2001placing}, is for use as a quality check for word embedding models. The aim of most embedding models is to capture a word's semantic relationships, such that words that are similar in meaning are placed close in the semantic space; foods with other foods, technical terms together and separated from the musical instruments, to give an example. However, the optimal performance of such a semantic space is judged by whether or not one wishes to capture similarity of words, or relatedness. It seems obvious that presenting \textit{cup} as a query reformulation for \textit{coffee} in information retrieval seems off, while presenting \textit{lamborghini} when searching for \textit{ferrari} can be completely acceptable. Inversely, in places where relatedness is needed, offering a \textit{cup} when one asks for a \textit{coffee} is correct. 

While the first word similarity datasets appeared for English, in the past few years we have seen  datasets for a range of different languages (see Section \ref{sec:relwork}). For Swedish, there exists one automatically-
created resource based on an association lexicon by \citet{fallgren2016towards}. However, there are to date no test sets that are (1) expertly-annotated, (2) comparable to other international test sets, and (3) annotated for both relatedness and similarity. 
And because we cannot know which motivation lies behind creating a vector space, and because both relatedness and similarity seem equally valid, we have opted to create \textit{SuperSim}. The SuperSim test set is a larger-scale similarity and relatedness set for Swedish, consisting of 1,301 words and 1,360 pairs rated by 5 expert annotators.
The pairs are based on SimLex-999 and WordSim353, and can be used to assess the performance of word embedding models, but also answer questions as to whether words are likely to be similar but not related. 

\section{Related Work}\label{sec:relwork}

Several works aim to provide test sets to assess the quality of word embedding models. Most of them tackle English \citep{rubenstein1965contextual,miller1991contextual,agirre2009study,bruni-etal-2012-distributional,hill-etal-2015-simlex}.
Russian, Italian and German are covered by \citet{leviant2015separated} who translated the pairs in WordSim353 and SimLex-999, and asked crowdworkers to judge them on a 0-10 scale. 
The SemEval-2017 Task 2 on Multilingual and Cross-lingual Semantic Word Similarity \citep{camacho-collados-etal-2017-semeval} provides pairs in 5 languages: English, Farsi, German, Italian and Spanish.
\citet{ercan-yildiz-2018-anlamver} provide 500 word pairs in Turkish annotated by 12 humans for both similarity and relatedness on a scale ranging from 0 to 10, while Finnish is covered in \citet{venekoski-vankka-2017-finnish}.
More recently, Multi-SimLex \citep{vulic2020multisimlex} provides annotations in Mandarin Chinese, Yue Chinese, Welsh, English, Estonian, Finnish, French, Hebrew, Polish, Russian, Spanish, Kiswahili, and Arabic, with open guidelines and encouragement to join in with more languages.\footnote{The website is updated with new annotations: \url{https://multisimlex.com/}.}

For Swedish, \citet{fallgren2016towards} harness the Swedish Association Lexicon SALDO \citep{borin2013saldo}, a large lexical-semantic resource that differs much from Wordnet \citep{fellbaum1998wordnet} insofar as it organises words mainly with the `association' relation.
The authors use SALDO's `supersenses' to adapt \citet{tsvetkov-etal-2016-correlation}'s QVEC-CCA intrinsic evaluation measure to Swedish.
Still on evaluating Swedish language models, \citet{adewumi2020exploring} propose an analogy test set built on the one proposed by \citet{mikolov2013efficient}, and evaluate common architectures on downstream tasks. 
The same authors further compare these architectures on models trained on different datasets (namely the Swedish Gigaword corpus \citep{eide2016swedish} and the Swedish Wikipedia) by focusing on Swedish and utilising their analogy test set \citep{adewumi2020corpora}.
Finally, for Swedish, SwedishGLUE/SuperLim\footnote{\href{https://spraakbanken.gu.se/projekt/superlim-en-svensk-testmangd-for-sprakmodeller}{\texttt{https://spraakbanken.gu.se/projekt/\\superlim-en-svensk-testmangd-for-\\sprakmodeller}}} \citep{Adesam-Yvonne2020-299130} is currently being developed as a benchmark suite for language models in Swedish, somewhat mirroring English counterparts \citep{wang2018glue,wang2019superglue}.

Whether similarity test sets actually allow to capture and evaluate lexical semantics is debatable \citep{faruqui-etal-2016-problems,schnabel-etal-2015-evaluation}. 
Nonetheless, they have the advantage of providing a straightforward way of optimising word embeddings (through hyper-parameter search, at the risk of overfitting), or to be used more creatively in other tasks \citep{dubossarsky-etal-2019-time} where ``quantifiable synonymy'' is required. Finally, task-specific evaluation (as recommended by \citep{faruqui-etal-2016-problems}) is, for languages other than English, more than often nonexistent -- making test sets such as the one presented in this work a good alternative.

Our dataset differs from previous work in the sense that it provides expert judgments for Swedish for both relatedness and similarity, and hence comprises two separate sets of judgments, as done by skilled annotators.\footnote{We have opted not to follow Multi-SimLex because (1) we want to have annotations for both relatedness and similarity, and (2) we have limited possibility to use platforms such as Amazon Mechanical Turk, and have thus resorted to using skilled annotators: to illustrate, we are bound to the hourly rate of 326 SEK (32.08 EUR). As a result the cost of annotating with 10 annotators is significantly higher, in particular if we want two separate sets of annotations.} A description of the procedure is available in Section~\ref{sec:desc}. 

\subsection{Relatedness and Similarity}

Our work heavily draws from \citet{hill-etal-2015-simlex}, who made a large distinction between relatedness and similarity. 
Indeed, the authors report that previous work such as \citet{agirre2009study} or \citet{bruni-etal-2012-distributional} do not consider relatedness and similarity to be different. Words like \emph{coffee} and \emph{cup}, to reuse the example by \citet{hill-etal-2015-simlex}, are obviously related (one is used to drink the other, they can both be found in a kitchen, etc.) but at the same time dissimilar (one is (...usually) a liquid and the other is a solid, one is ingested and not the other, etc.). 

All pairs in SuperSim are independently judged for similarity and relatedness. To explain the concept of similarity to annotators, we have reused the approach of \citet{hill-etal-2015-simlex} who introduced it via the idea of synonymy, and in contrast to association: ``In contrast, although the following word pairs are related, they are not very similar. The words represent entirely different types of things.'' They further give the example of ``car / tyre.''
We use this definition embedded in the SimLex-999 guidelines to define relatedness according to the following: ``In Task 2, we also ask that you rate the same word pairs for their relatedness. For this task, consider the inverse of similarity: \textit{car} and \textit{tyre} are related even if they are not synonyms. However, synonyms are also related."

\section{Dataset description}
\label{sec:desc}

While the WordSim353 pairs were chosen for use in information retrieval and to some extent mix similarity and relatedness, the original SimLex-999 pairs were chosen with more care. They were meant to measure the ability of different models to capture similarity as opposed to association, contain words from different part-of-speech (nouns, verbs, and adjectives), and represent different concreteness levels. Despite the risks of losing some intended effect in translation, we opted to base SuperSim on both of these resources rather than start from scratch.

\subsection{Methodology}

We machine-translated all words in WordSim353 and SimLex-999 to Swedish. The translations were manually checked by a semanticist who is a native speaker of Swedish, holds an MA in linguistics, and is currently working towards obtaining a PhD in linguistics. The semanticist was presented a list of words, out of context, decoupled from the pairs they were parts of. 
Where needed, translations were corrected.
Pairs were reconstructed according to the original datasets, except for the few cases where the translation process would create duplicates. In a few cases where one single translation was not obvious -- i.e. cases where either Google Translate or the semanticist would output two (equally likely) possible Swedish translations for the same English word --, two pairs were constructed: one with each possible translation. For example, the presence of `drug' led to pairs with both the \emph{läkemedel} (a medical drug aimed at treating pathologies) and \emph{drog} (a narcotic or stimulant substance, usually illicit) translations.

We selected 5 annotators (4F/1M) who are native speakers of Swedish and all have experience working with annotation tasks.  One of the annotators was the same person who manually checked the correctness of the translations. 
The other 4 annotators can be described as follows:
\begin{itemize}
    \item holds an MA in linguistics and has experience in lexicography, 
    \item holds an MA in linguistics,
    \item holds BAs in linguistics and Spanish and is studying for an MSc in language technology,
    \item holds a BA in linguistics and has extensive work experience with different language-related tasks such as translation and NLP (on top of annotation).
\end{itemize}

Annotators were each given (i) the original SimLex-999 annotation instructions containing examples illustrating the difference between relatedness and similarity; (ii) one file for the relatedness scores; and (iii) one file for the similarity scores. They were instructed to complete the annotation for similarity before moving on to relatedness, and complied. The annotation took place, and was monitored, on Google Sheets. Annotators did not have access to each others' sheets, nor were they aware of who the other annotators were. 

To allow for a finer granularity as well as to echo previous work, 
annotators were tasked with assigning scores on a 0-10 scale, rather than 1-6 as in SimLex-999. Unlike the procedure for Simlex, where sliders were given (and hence the annotators could choose real values), our annotators assigned discrete values between 0--10. This procedure resulted in pairs with the same score, and thus many rank ties.

\subsection{SuperSim stats}
The entire SuperSim consists of 1,360 pairs. Out of these, 351 pairs stem from WordSim353 and 997 pairs from SimLex-999.  Pairs where both words translate into one in Swedish are removed from the SimLex-999 and WordSim353 subsets, thus resulting in fewer pairs than the original datasets: for example, `engine' and `motor' are both translated as \textit{motor} and therefore the `motor' -- `engine' pair is removed. 
The SuperSim set consists of both sets, as well as of a set of additional pairs where multiple translations were used (see the \textit{läkemedel} and \textit{drog} example above). The full set of 1,360 pairs is annotated for both similarity and relatedness separately, resulting in a total of 2 * 1,360 gold scores, and thus 13,600 individual judgments. An example of relatedness judgments for two pairs is available in table form in Table~\ref{tbl:rel-ex}.

\begin{table*}[t]
\centering
\small
\caption{Example of relatedness judgments on pairs \textit{flicka-barn} `girl-child' and \textit{skola-mitten} `school-centre.'}
\label{tbl:rel-ex}
\begin{tabular}{llrrrrrr}
\toprule
Word 1  &  Word 2  &  Anno 1  &  Anno 2  &  Anno 3  &  Anno 4  &  Anno 5  &  Average\\
\midrule
flicka & barn & 10 & 10 & 10 & 8 & 10 & 9.6\\
skola & mitten & 1 & 0 & 0 & 0 & 0 & 0.2\\
\bottomrule
\end{tabular}
\end{table*}

We release two tab-separated files (one for relatedness, one for similarity) containing judgments from all annotators as well as the mean gold score. We additionally release all baseline models, code, and pre-processed data where permissible. The data is freely available for download at \url{https://zenodo.org/record/4660084}.

\subsection{Intra-rater agreement}
For quality control, annotation files contained a total of 69 randomly sampled duplicate pairs, in addition to the 1,360 true pairs.\footnote{SuperSim includes the values for the first seen annotation of a duplicate pair. To illustrate: if a control pair was annotated first to have a score of 3 and then to have a score of 6, the first score of 3 is kept.} These duplicates allowed us to calculate every annotator's consistency, and to judge how difficult each task was in practice. Table~\ref{tbl:anno-dis} illustrates the consistency of every annotator in the similarity and relatedness tasks for our 69 control pairs. `Disagreement' indicates two different values for any given pair and `hard disagreement' two values with an absolute difference higher than 2 (on the scale of 0--10). On average, the hard disagreements differed by 4.3 points for relatedness, and by 3.0 for similarity, and there were more disagreements (both kinds) for relatedness, indicating that for humans, relatedness is the harder task.   
In addition, we indicate the computed self-agreement score (Krippendorff's alpha, \citealt{krippendorff2018content}) for every annotator for both tasks. Despite annotators disagreeing somewhat with themselves, Krippendorff's alpha indicates they annotated word pairs consistently. 

Out of the 69 control pairs, 4 were inconsistently annotated by four annotators for similarity, while 12 pairs were inconsistently annotated by four or more annotators for relatedness: 3 by all five annotators, and 9 by four. The three ``hardest'' pairs to annotate for relatedness are \emph{lycklig-arg} `happy-angry,' \emph{sommar-natur} `summer-nature,' \emph{tillkännagivande-varning} `announcement-warning.' 


\begin{table*}[t]
\centering
\small
\caption{Number of control word-pairs with annotator self-disagreements. `Disagreem.' = different values between two annotations for a given pair (0-10 scale), `hard disagreem.' = difference $>$ 2 between values between two annotations for a given pair (0-10 scale), $\alpha$ = Krippendorff's alpha. Total number of control pairs is 69, percentages follow absolute counts in parentheses.}
\label{tbl:anno-dis}
\begin{tabular}{l|rrr|rrr}
\toprule
& \multicolumn{5}{c}{Consistency of judgments}\\
&  \multicolumn{2}{c}{Similarity} & \multicolumn{3}{c}{Relatedness} \\
& \# disagreem. (\%) & \# hard disagreem. (\%) & $\alpha$ & \# disagreem. (\%) & \# hard disagreem. (\%) & $\alpha$\\
\midrule
Anno 1 & 17 (25\%) & 5 (7\%) & 0.83 & 20 (29\%) & 10 (14\%) & 0.89 \\
Anno 2 & 1 (1\%) & 1 (1\%) & 0.99 & 26 (38\%) & 11 (16\%) & 0.86 \\
Anno 3 & 21 (30\%) & 6 (9\%) & 0.94 & 24 (35\%) & 9 (13\%) & 0.87 \\
Anno 4 & 10 (14\%) & 0 (0\%) & 0.96 & 18 (26\%) & 4 (8\%) & 0.96 \\
Anno 5 & 29 (42\%) & 3 (4\%) & 0.89 & 28 (41\%) & 7 (10\%) & 0.89 \\
\bottomrule
\end{tabular}
\end{table*}

\subsection{Inter-rater agreement}
Following  \citet{hill-etal-2015-simlex}, we use the average Spearman's $\rho$ for measuring inter-rater agreement by taking the average of pairwise Spearman's $\rho$ correlations between the ratings of all respondents.\footnote{We use the \texttt{scipy.stats.mstats spearmanr} \cite{2020SciPy-NMeth} implementation with rank ties.} 
For the original SimLex-999, overall agreement was $\rho$ = 0.67 as compared to WordSim353 where $\rho$ = 0.61 using the same method. 
Spearman's $\rho$ for our similarity rankings is 0.67. In addition, we have a Spearman's $\rho$ for our relatedness rankings of 0.73.\footnote{These results are opposing those of the disagreements which indicate that similarity is easier than relatedness for our annotators. We postulate that this can be due to the many rank ties we have in the similarity testset (where many pairs have 0 similarity). If we use the Pearson's $\rho$, we get values of $\rho =$ 0.722 for relatedness, and $\rho =$ 0.715 for similarity bringing the two tasks much closer. }
It is unclear how the background of our annotators affects the quality of their annotation. 
In another semantic annotation study, although on historical data, \citet{schlechtweg-etal-2018-diachronic} show a larger agreement between annotators sharing a background in historical linguistics than between a historical linguist and a `non-expert' native speaker. It is, however, fully possible that the linguistic expertise of the annotators affects the similarity and relatedness judgments in a negative way. We leave this investigation for further work.

\section{Model evaluation}
\label{sec:baseline-eval}

To provide a baseline for evaluation of embedding models on SuperSim, we trained three different models on two separate datasets.

\begin{table*}[t]
\centering
\small
\caption{Evaluation of models trained on the Swedish Gigaword corpus. WordSim353 and SimLex-999 are subsets of the SuperSim. Best results for each ``test set - task'' combination are bolded.}
\label{tbl:baselines-models-gigaword}
\begin{tabular}{llrrr}
\toprule
\multirow{2}{*}{Model} &  \multirow{2}{*}{Test set} & Spearman's $\rho$ & Spearman's $\rho$ & \multirow{2}{*}{Included pairs}  \\
& & relatedness & similarity &\\
\midrule
\multirow{3}{*}{Word2Vec}  & SuperSim  & 0.539& 0.496 & 1,255\\
 & WordSim353 pairs & \textbf{0.560} & 0.453 & 325\\
 & SimLex-999 pairs & 0.499& 0.436& 923\\
\midrule
\multirow{3}{*}{fastText}  & SuperSim & \textbf{0.550}& \textbf{0.528}& 1,297\\
 & WordSim353 pairs &  0.547& \textbf{0.477} &347\\
  & SimLex-999 pairs & \textbf{0.520}& \textbf{0.471}& 942\\
\midrule
 \multirow{3}{*}{GloVe} & SuperSim & 0.548& 0.499&1,255\\
  & WordSim353 pairs & 0.546& 0.435 & 325\\
  & SimLex-999 pairs & 0.516 & 0.448 & 923\\
\bottomrule
\end{tabular}
\end{table*}

\begin{table*}[t]
\centering
\small
\caption{Evaluation of models trained on the Swedish Wikipedia. WordSim353 and SimLex-999 are subsets of the SuperSim. Best results for each ``test set - task'' combination are bolded.}
\label{tbl:baselines-models-wiki}
\begin{tabular}{llrrr}
\toprule
\multirow{2}{*}{Model} &  \multirow{2}{*}{Test set} & Spearman's $\rho$ & Spearman's $\rho$ & \multirow{2}{*}{Included pairs} \\
& & relatedness & similarity& \\
\midrule
\multirow{3}{*}{Word2Vec}  & SuperSim & 0.410 & 0.410 & 1,197\\
 & WordSim353 pairs & 0.469& 0.415 & 315\\
 & SimLex-999 pairs & 0.352 & 0.337 & 876\\
\midrule
\multirow{3}{*}{fastText}  & SuperSim & 0.349 & 0.365 & 1,297\\
  & WordSim353 pairs &  0.339 & 0.334 & 347\\
  & SimLex-999 pairs & 0.322 & 0.311 & 942\\
\midrule
\multirow{3}{*}{GloVe}  & SuperSim & \textbf{0.467} & \textbf{0.440} & 1,197\\
  & WordSim353 pairs & \textbf{0.524}& \textbf{0.429} & 315\\
 & SimLex-999 pairs & \textbf{0.418} & \textbf{0.375} &876\\
\bottomrule
\end{tabular}
\end{table*}

\subsection{Baseline Models}
We chose three standard models, Word2Vec \citep{mikolov2013efficient}, fastText \citep{bojanowski2017enriching}, and GloVe \citep{pennington2014glove}. 
Word2Vec and fastText models are trained with gensim \citep{rehurek_lrec} while the GloVe embeddings are  trained using the official C implementation provided by \citet{pennington2014glove}.\footnote{Tests were also made using the Python implementation available at \url{https://github.com/maciejkula/glove-python}, with similar performance.}

\subsection{Training data}
We use two datasets. 
The largest of the two comprises the Swedish Culturomics Gigaword corpus \citep{eide2016swedish}, which contains a billion words\footnote{1,015,635,151 tokens in 59,736,642 sentences, to be precise.} in Swedish from different sources including fiction, government, news, science, and social media. 
The second dataset is a recent Swedish Wikipedia dump with a total of 696,500,782 tokens.\footnote{Available at \url{https://dumps.wikimedia.org/svwiki/20201020/svwiki-20201020-pages-articles.xml.bz2}.} 

While the Swedish Gigaword corpus contains text from the Swedish Wikipedia, \citet{eide2016swedish} precise that about 150M tokens out of the 1G in Gigaword (14.9\%) stem from the Swedish Wikipedia. In that respect, there is an overlap in terms of content in our baseline corpora. However, as the Swedish Wikipedia has grown extensively over the years and only a sub-part of it was used in in \citet{eide2016swedish}, the overlap is small and we thus have opted to also use the Gigaword corpus as it is substantially larger and contains other genres of text.

The Wikipedia dump was processed with a version of the Perl script released by Matt Mahoney\footnote{The script is available at \url{ http://mattmahoney.net/dc/textdata.html}. It effectively only keeps what should be displayed in a web browser  and removes tables but keeps image captions, while links are converted to normal text. Characters are lowercased.} modified to account for specific non-ASCII characters (\texttt{äåöé}) and to transform digits to their Swedish written form (eg: 2 $\rightarrow$ \textit{två}).\footnote{`1', which can be either \emph{en} or \emph{ett} in Swedish, was replaced by `ett' every time.}

All baseline models are trained on lowercased tokens with default hyperparameters.\footnote{Except for $sg = 1$, $min\_count = 100$ and $seed = 1830$.}

\subsection{Results}
An overview of the performance of the three baseline models is available in Table~\ref{tbl:baselines-models-gigaword} and Table~\ref{tbl:baselines-models-wiki}.  In both tables we show model performance on similarity and relatedness judgments. We split the results into three sets, one for the entire SuperSim, and two for its subsets: WordSim353  
and SimLex-999. For each model and dataset, we present Spearman's rank correlation $\rho$ between the ranking produced by the model compared to the gold ranking in each testset (relatedness and similarity). As fastText uses subword information to build vectors, it deals better with out-of-vocabulary words, hence the higher number of pairs included in the evaluation.

To provide a partial reference point, \citet{hill-etal-2015-simlex} report,  for Word2Vec trained on English Wikipedia, $\rho$ scores of 0.655 on WordSim353, and 0.414 on SimLex-999. 

From the results in Table \ref{tbl:baselines-models-gigaword} and \ref{tbl:baselines-models-wiki}, it appears that 
fastText is the most impacted by the size of the training data, as its performance when trained on the smaller Wikipedia corpus is `much' lower than on the larger Gigaword: 0.349 vs 0.550 for SuperSim relatedness and 0.365 vs 0.528 for Supersim similarity -- both tasks where fastText actually performs best on Gigawords out of the three models tested.
We find that all models perform better when trained on Gigaword as compared to Wikipedia. Contrary to results on the analogy task reported by \citet{adewumi2020corpora}, our experiments on SuperSim seem to confirm the usual trope that training on more data indeed leads to \textbf{overall} better embeddings, as the higher scores, in terms of absolute numbers, are all from models trained on the larger Gigaword corpus. 
Nonetheless, the discrepancy between our results and theirs might be due to a range of factors, including pre-processing and hyperparameter tuning (which we did not do).\footnote{The effect of the benefits of more training data is confounded with the broader genre definitions in Gigaword that could be an indication of the advantage of including e.g., fiction and social media text in defining for example emotions. We leave a detailed investigation into this for future work.}

Note that for similarity, Word2Vec trained on Gigaword performs slightly better on the translated SimLex-999 pairs (0.436) than Word2Vec does on English SimLex-999 (0.414) but substantially lower for WordSim (0.436 vs 0.655) \citep{hill-etal-2015-simlex}. We make the comparison for Gigaword, rather than Wikipedia because of the comparable size, rather than the genre. This effect could be due to different pre-processing and model parameters used, but it could also be an effect of the multiple ties present in our test set. 
We do, however, consistently confirm the original conclusion:\textbf{ SimLex-999 seems harder for the models than WordSim353}. 

GloVe is the clear winner on the smaller Wikipedia dataset, where it outperforms the other two models for all test sets, and is on par with Word2Vec for Gigaword.  

Overall, our results indicate that  \textbf{for the tested models relatedness is an easier task than similarity}: every model -- aside from fastText on SuperSim -- performs better (or equally well) on relatedness on the whole test set, as well as on its subparts, compared to similarity. 

\section{Conclusions and future work}
In this paper, we presented SuperSim, a Swedish similarity and relatedness test set made of new judgments of the translated pairs of both SimLex-999 and WordSim353. All pairs have been rated by five expert annotators, independently for both similarity and relatedness. 
Our inter-annotator agreements mimic those of the original test sets, but also indicate that similarity  is an easier task to rate than relatedness, while our intra-rater agreements on 69 control pairs indicate that the annotation is reasonably consistent. 
\balance 

To provide a baseline for model performance, we trained three different models, namely Word2Vec, fastText and GloVe, on two separate Swedish datasets. The first comprises a general purpose dataset, namely the The Swedish Culturomics Gigaword Corpus with different genres of text spanning 1950-2015. The second comprises a recent Swedish Wikipedia dump. On the Gigaword corpus, we find that fastText is best at capturing both relatedness and similarity while for Wikipedia, GloVe performs the best.

Finally, to answer the question posed in the introduction: it is common to have words that are highly related, but not similar. To give a few examples, these are pairs with relatedness 10 and similarity 0: \emph{bil-motorväg} `car-highway,' \emph{datum-kalender} `date-calendar,' \emph{ord-ordbok} `word-dictionary,' \emph{skola-betyg} `school-grade,' and \emph{tennis-racket} `tennis-racket.' 

The opposite however, does not hold. Only four pairs have a similarity score higher than the relatedness score, and in all cases the difference is smaller than 0.6: \emph{bli-verka} `become-seem,' \emph{rör-cigarr} `pipe-cigarr,' \emph{ståltråd-sladd} `wire-cord,' \emph{tillägna sig-skaffa sig} `get-acquire.' 

For future work, the SuperSim testset can be improved both in terms of added annotations (more annotators), and with respect to more fine-grained judgements (real values in contrast to discrete ones currently used) to reduce the number of rank ties.

\section{Acknowledgments}
We would like to thank Tosin P. Adewumi, Lidia Pivovarova, Elaine Zosa, Sasha (Aleksandrs) Berdicevskis, Lars Borin, Erika Wauthia, Haim Dubossarsky, Stian R{\o}dven-Eide as well as the anonymous reviewers for their insightful comments. 
This work has been funded in part by the project \textit{Towards Computational Lexical Semantic Change Detection} supported by the Swedish Research Council (2019--2022; dnr 2018-01184), and \emph{Nationella Språkbanken} (the Swedish National Language Bank), jointly funded  by  the Swedish Research Council (2018--2024; dnr 2017-00626) and its ten partner institutions.

\bibliographystyle{acl_natbib}
\bibliography{additionalbib,anthology}

\end{document}